\documentclass[conference]{IEEEtran}
\IEEEoverridecommandlockouts
\usepackage{times}
\usepackage[T1]{fontenc}

\usepackage{cite}
\usepackage{amsmath,amssymb,amsfonts}
\usepackage{subfigure}
\usepackage{array}
\usepackage{graphicx}
\usepackage{textcomp}
\usepackage{gensymb}
\usepackage{xcolor}
\def\BibTeX{{\rm B\kern-.05em{\sc i\kern-.025em b}\kern-.08em
		T\kern-.1667em\lower.7ex\hbox{E}\kern-.125emX}}
\usepackage{algpseudocode}
\usepackage[ruled,vlined]{algorithm2e}
\usepackage{url}

\usepackage{amsthm} 
\newtheorem{lemma}{Lemma} 
\newtheorem{theorem}{Theorem} 

\usepackage{lipsum}
\usepackage{fancyhdr}
\begin{document}
	

\title{TMAV: \textbf{T}emporal \textbf{M}otionless \textbf{A}nalysis of \textbf{V}ideo using CNN in MPSoC\\
}
	
	\author{Somdip Dey,~\IEEEmembership{Student Member,~IEEE,}
		Amit K. Singh,~\IEEEmembership{Member,~IEEE,}
		Dilip K. Prasad,~\IEEEmembership{Member,~IEEE,}\\
		and~Klaus D. McDonald-Maier,~\IEEEmembership{Senior Member,~IEEE}
		\IEEEcompsocitemizethanks{\IEEEcompsocthanksitem S. Dey, A. K. Singh and K. D. McDonald-Maier are with the Embedded and Intelligent Systems Laboratory, University of Essex, UK.\protect\\
			E-mail: somdip.dey@essex.ac.uk
			\IEEEcompsocthanksitem D. K. Prasad is with the Arctic University of Norway, Norway.}
		\thanks{Copyright 2018 by the author(s).}}
	
	\maketitle

	\begin{abstract}
		Analyzing video for traffic categorization is an important pillar of Intelligent Transport Systems. However, it is difficult to analyze and predict traffic based on image frames because  the representation of each frame may vary significantly within a short time period. This also would inaccurately represent the traffic over a longer period of time such as the case of video. We propose a novel bio-inspired methodology that integrates analysis of the previous image frames of the video to represent the analysis of the current image frame, the same way a human being analyzes the current situation based on past experience. In our proposed methodology, called \textit{IRON-MAN} (\textbf{I}ntegrated \textbf{R}ational prediction and \textbf{M}otionless \textbf{AN}alysis), we utilize Bayesian update on top of the individual image frame analysis in the videos and this has resulted in  highly accurate prediction of Temporal Motionless Analysis of the Videos (\textit{TMAV}) for most of the chosen test cases. The proposed approach could be used for \textit{TMAV} using Convolutional Neural Network (CNN) for applications where the number of objects in an image is the deciding factor for prediction and results also show that our proposed approach outperforms the state-of-the-art for the chosen test case. We also introduce a new metric named, \textit{Energy Consumption per Training Image} (\textit{ECTI}). Since, different CNN based models have different training capability and computing resource utilization, some of the models are more suitable for embedded device implementation than the others, and \textit{ECTI} metric is useful to assess the suitability of using a CNN model in multi-processor systems-on-chips (MPSoCs) with a focus on energy consumption and reliability in terms of lifespan of the embedded device using these MPSoCs.
		
	\end{abstract}

	\section{Introduction and Motivation}
	\IEEEPARstart{R}{ecently} there has been a huge increase in utilizing Convolutional Neural Networks (CNNs) \cite{chakradhar2010dynamically, chen2014big, vgg2014} to solve several real-life challenges such as traffic categorization \cite{luo2015traffic}, \cite{kalliatakis2017detection}, weather forecasting \cite{zhang1998forecasting, grecu1999detection}, etc due to its high prediction accuracy/categorization in the aforementioned target applications. One particular case for harnessing the efficacy of visual CNN based prediction model is Intelligent Transportation Systems (ITS), which is becoming an important pillar in the modern ``smart city'' framework.
	
	\begin{figure}[!h]
		\centering
		\subfigure[Frame predicted as \textit{Light} traffic category]{\includegraphics[width=4.1cm]{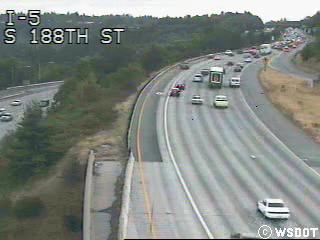}}
		\subfigure[Frame predicted as \textit{Heavy} traffic category]{\includegraphics[width=4.1cm]{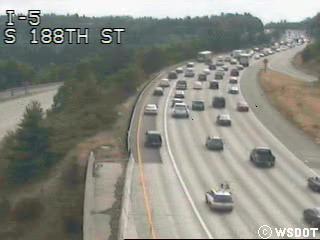}}
		\caption{Frames (Images) from the same \textit{Light} traffic category of UCSD dataset \cite{datasetChan2005classification} and associated prediction by a trained CNN model \cite{dey2018mat}}
		\label{motivation1}
	\end{figure} 
	
	Traffic load categorization is challenging given the increase in vehicles on road. Some of the popular ways of monitoring and categorizing traffic load from videos include vehicle based assess method \cite{asmaa2013road, hu2012real, riaz2013traffic, luo2015traffic} and holistic approach \cite{derpanis2011classification, porikli2004traffic, ess2009segmentation}. In vehicle based assess methodologies, either vehicles are first localized on the road with a background subtraction method \cite{jung2001content, andrews2013highway, asmaa2013road} or the vehicles are localized with moving feature keypoints \cite{kamijo2000traffic, hu2012real, riaz2013traffic}. Whereas, in holistic approach, a macroscopic analysis of traffic flow is understood through global representation of a scene, which is obtained by accounting for spatio-temporal features except tracking using background subtraction and moving feature keypoints \cite{chan2005classification, porikli2004traffic, derpanis2011classification}. 
	
	In recent times, there has also been emergence of several methods capable of monitoring and analyzing traffic using motionless analysis of videos \cite{luo2015traffic, luo2016traffic, dey2018mat}, where videos of traffic are broken into frames instead, and the frames are analyzed for further computation or prediction. The main motivation to utilize methodologies consisting of motionless analysis of video is that it is difficult to stream high-frame rate videos gathered by a large network of interconnected cameras due to bandwidth limitation. Hence, streaming low-frame rate videos on these camera networks is very common. In many cases, it is challenging to stream more than 2 frames per second due to the limited bandwidth of the network when these cameras stream over a WIFI network \cite{luo2015traffic, dey2018mat}. Moreover, to analyze video in real-time without motion features over a WIFI network is difficult due to communication bandwidth constraint and hence, it is better to analyze the image fames on the camera enabled embedded device itself \cite{dey2018mat} instead of relying on a server system over th WIFI network. Another motivation to devise such approaches in embedded devices is the afford-ability of such devices instead of employing powerful server systems used for analysis purposes\cite{dey2018mat}. Therefore, the approach of analyzing videos without motion features on the embedded device is not just beneficial for categorizing traffic load but could be extended to several computer vision based real-world application that requires analysis of low-frame rate videos. Although motionless analysis of videos have their own benefits, they also come with limitations described with the following observations.

	\noindent\textbf{Observation 1}: Although several implementations of such methods were able to achieve high prediction accuracy on known dataset \cite{luo2015traffic, luo2016traffic, dey2018mat}, but in some test cases the analysis were not accurate at all. The reason for failed prediction/analysis is that in some cases it is difficult to predict the label of an image frame from a video if the ground truth of the image is overlapping with several other categories (labels) instead of falling under one category. For example, in the dataset of traffic released by UCSD \cite{luo2015traffic, datasetChan2005classification}, which consists of light, heavy and traffic jam categories, two frames (images) belonging to the same category of video are predicted differently by the CNN model \cite{dey2018mat}. The reason for such behavior is that the CNN predicts the label and probability of it occurring on the instantaneous image frame. In Fig. \ref{motivation1}, we notice that a trained CNN model \cite{dey2018mat} with an overall prediction accuracy of 81.25\% predicted the wrong label for a frame, which falls under \textit{Light} category but was instead predicted as \textit{Heavy}. However, both the fames belong to the same video under the \textit{Light} category. If the ground truth\footnote{Ground truth of the image frame in this case is the information gained through empirical evidence as opposed to the inference made by the CNN model.} of the two images ($(a)~\&~(b)$) are compared then it is justifiable that the prediction by the CNN is in fact accurate due to the fact that the traffic projected in Fig. \ref{motivation1}.$(b)$ is more congested than the traffic projected in \ref{motivation1}.$(a)$. In reality the analysis of each image frame of the video should also portray the overall analysis\footnote{Here, overall analysis of video means the analysis of the video as a whole as opposed to the analysis of each image frame of the video.} for the video instead of the frame itself in order to convey the temporal prediction. Since, each individual image frame of a video could lead to different analysis result (prediction/label), the temporal prediction is the prediction analysis over time. This limitation is due to the fact that the trained CNN model only predicts the label or analyzes the current image frame without taking past image frames into consideration. Therefore, the challenge is to analyze and predict videos just from the image frames without motion features of the video and yet give accurate temporal prediction results for the video as well. Although there have been some recent studies, which focused on future predictions of motion in ego-centric videos \cite{yagi2017future, lai2018people}, such as predicting the future position of a person based on the current image frame, but no study to our best knowledge have tried to predict the scene\footnote{Scene is a place where a human being could navigate or can act within.} \cite{zhou2014learning, xiao2010sun} of a video from image frames taking predictions from immediate previous frames into account to provide a more holistic analysis over a time period. Hence, we call such an analysis as \textit{\textbf{T}emporal \textbf{M}otionless \textbf{A}nalysis of \textbf{V}ideo (TMAV)} and several target applications of CNN such as traffic categorization require such kind of analysis in comparison with traditional one \cite{luo2015traffic, luo2018traffic, dey2018mat}.
	
	\noindent\textbf{Observation 2}: In the study \cite{dey2018mat}, Dey et al. proposed a methodology to implement a CNN which is trained on a configurable embedded device and it was utilized to categorize traffic on the same device. Although this study gave a novel approach on analyzing traffic using video cameras in low bandwidth network without the need to communicate the image frames over the WIFI network, the study also had energy consumption and device lifespan reliability issues (shown later in the section). When the CNN model is implemented and trained on a multi-processor system-on-a-chip (MPSoC) such as Odroid XU4 \cite{odroidxu4}, the maximum temperature of the CPU reached 93.72\degree C on an average and the power consumption peaked at 10.63 Watt on an average during the training period. Reaching a high operating temperature for a long period of time is an important factor in the reduction of lifespan of the device. In some studies \cite{chantem2008temperature, coskun2007temperature, zhou2016thermal} it has been found that an increase in the operating temperature by 10-15\degree~centigrades could reduce the lifespan of the device by 2$\times$. An increase in operating temperature of the device could be both temporal and spatial \cite{coskun2007temperature}, where increase in temperature over a time period is known as an increase in temporal thermal gradient and an increase in operating temperature with respect to space is called an increase in spatial thermal gradient. Additionally, an increase in energy consumption on embedded devices is harmful for battery operation itself \cite{panigrahi2001battery, lahiri2002battery}, especially in low-power embedded devices. Thus it is important to design not just an energy-efficient CNN model, which is able to achieve desired analysis of video without motion features, but at the same time does not affect the lifespan reliability of the system adversely.
	
	In order to overcome the limitations of the existing approaches we propose \textbf{IRON-MAN}: \textbf{I}ntegrated \textbf{R}ati\textbf{ON}al prediction and \textbf{M}otionless \textbf{AN}alysis of videos using CNN, which is capable of performing $TMAV$, in MPSoCs. To this end, this paper makes the following contributions:
	\begin{enumerate}
		\item An energy efficient prediction methodology (\textit{IRON-MAN}) which integrates predictions of previous frames of a video to predict the current frame and hence analyze the video without using motion features.
		\item A new metric named \textit{Energy Consumption per Training Image} ($ECTI$), which will enable the choice of suitable CNN model for real-world applications on embedded devices keeping energy-efficiency in mind.
		\item Validation of the proposed approach on a real hardware platform, the Odroid-XU4 \cite{odroidxu4}.
		\item Comparative study of energy consumption using $ECTI$ metric and temporal thermal gradient on the device while training the model on the device and off the device\footnote{Training a CNN model could be performed both on the embedded device or off the device. During `off the device training' it could be performed on a more resourceful device with higher computing resources (CPU/GPU/Memory) and then the weights and parameters of the trained CNN model are saved and transfered to the embedded system for prediction/analysis.}.
		\item Effect on lifespan of the device utilizing CNN based approaches on such platforms
	\end{enumerate} 

\section{Related Work}\label{relatedwork}
Majority of traffic analysis and categorization approaches before 2015 were mostly performed using the following methodologies:

\begin{itemize}
	\item Vehicle based methodologies where either vehicles are first localized on the road with a background subtraction method \cite{jung2001content, andrews2013highway, asmaa2013road} or the vehicles are localized with moving feature keypoints \cite{ hu2012real, riaz2013traffic}. In these methodologies the resulting tracks are concatenated together to identify key features of traffic such as traffic lanes, average traffic speed, average traffic density, etc.
	\item A holistic approach, where a macroscopic analysis of traffic flow is understood through global representation of a scene, which is obtained by accounting for spatio-temporal features except tracking using background subtraction and moving feature keypoints \cite{chan2005classification, porikli2004traffic, derpanis2011classification}.
\end{itemize}

Although the aforementioned methodologies are highly effective to analyze traffic, the biggest limiting factor is the cost of sophisticated camera-network involved and the requirement for high-frame-rate videos to compute reliable motion features. To break away from this trend of traffic analysis, in 2015 Luo et al. \cite{luo2015traffic} proposed a methodology to use various image processing and CNN based approaches to analyze traffic without moving features. In this paper the authors used four different visual descriptors such as bag of visual words (BOVW), Vector of Locally Aggregated Descriptors (VLAD), improved Fisher Vector (IFV) and Locality-constrained Linear Coding (LLC), and have also used pre-trained deep CNN models such as Caffe and VGG to analyze traffic and predict categorization of the same. The approach taken by Luo et al. to use popular image processing and CNN methods to classify traffic is novel and solves the low-frame-rate video streaming issue. In another extended paper published by Luo et al. \cite{luo2016traffic}, the researchers have used SegCNN and RegCNN to analyze and classify traffic. In both the aforementioned papers the authors are training and classifying traffic images after the video frames are transferred to the server from the interconnected camera network. But installing and implementing such hardware infrastructure to analyze traffic in developing countries is a challenging issue\cite{mokhaareview} due to the associated implementation costs \cite{dey2018mat}. In another study \cite{luo2018traffic}, Luo et al. utilizes image segmentation, Deep Learning \cite{lecun2015deep} and analysis techniques with CNN to learn features from the traffic scene and categorize without motion features. The traffic analysis methodologies proposed in \cite{luo2015traffic, luo2016traffic, luo2018traffic} are not capable of being implemented on low powered embedded devices during the training phase because such methodologies utilize a lot of computing resources due to the computational cost involved. In \cite{luo2015traffic, luo2016traffic, luo2018traffic}, training of the CNN model has to be performed on a powerful computer with a lot of computing resources and then use the trained CNN on the device for prediction/analysis, which makes the methodologies inflexible to new training dataset. Every time the models require to be trained on a powerful computing device and only be used on the embedded device for prediction/analysis.

In the study \cite{dey2018mat}, Dey et al. proposed a traffic categorization methodology using CNN where the training could be performed on the device mimicking a human being's ability to learn from its surroundings and makes such an approach more flexible as opposed to \cite{luo2015traffic, luo2016traffic, luo2018traffic}. However, neither of the studies in \cite{luo2015traffic, luo2016traffic, luo2018traffic, dey2018mat} are capable of performing temporal analysis of video, where a holistic analysis of video could be provided instead of analyzing each image frame.

Other state-of-the-art methodologies include detecting \& counting the numbers of cars and computing traffic density based on that using CNN-based vehicle detectors with high accuracy at near real time \cite{zhang2015integrated, cai2016unified, zhou2016image}. Although this way of detecting traffic density could still be classified as a vehicle based approach and has become popular in recent times but there are associated challenges with these methods as follows:
\begin{itemize}
	\item Training and test data should belong to the same dataset taken from the same camera with same configuration and hence require consistency in training.
	\item Cars detected need to be within a particular range or scope of the image and these methodologies fail to detect cars, which are far away in the images captured.
	\item These methodologies performed poorly if the captured images were occluded, especially in case of heavy traffic \& jam.
\end{itemize}


Compared to all the aforementioned works, we propose an easy to train CNN model, which do not require a lot of images in the training dataset, with combination of transfer learning\footnote{Learning achieved by taking the convolutional base of a pre-trained network, running the new data of 4 traffic categories through it and training a new randomly initialized classifier}, continuous learning\footnote{Learning achieved by re-training the classifier with wrong predictions till operating period of the system} and Bayesian update\footnote{Bayesian update is the ability to update the probability for a hypothesis as more new information or evidence becomes available.} capabilities on the MPSoC.
	
	\section{Proposed Methodology: IRON-MAN}\label{mainmethod}
\begin{figure}[htbp]
		\centerline{\includegraphics[width=9cm,height=7.9cm]{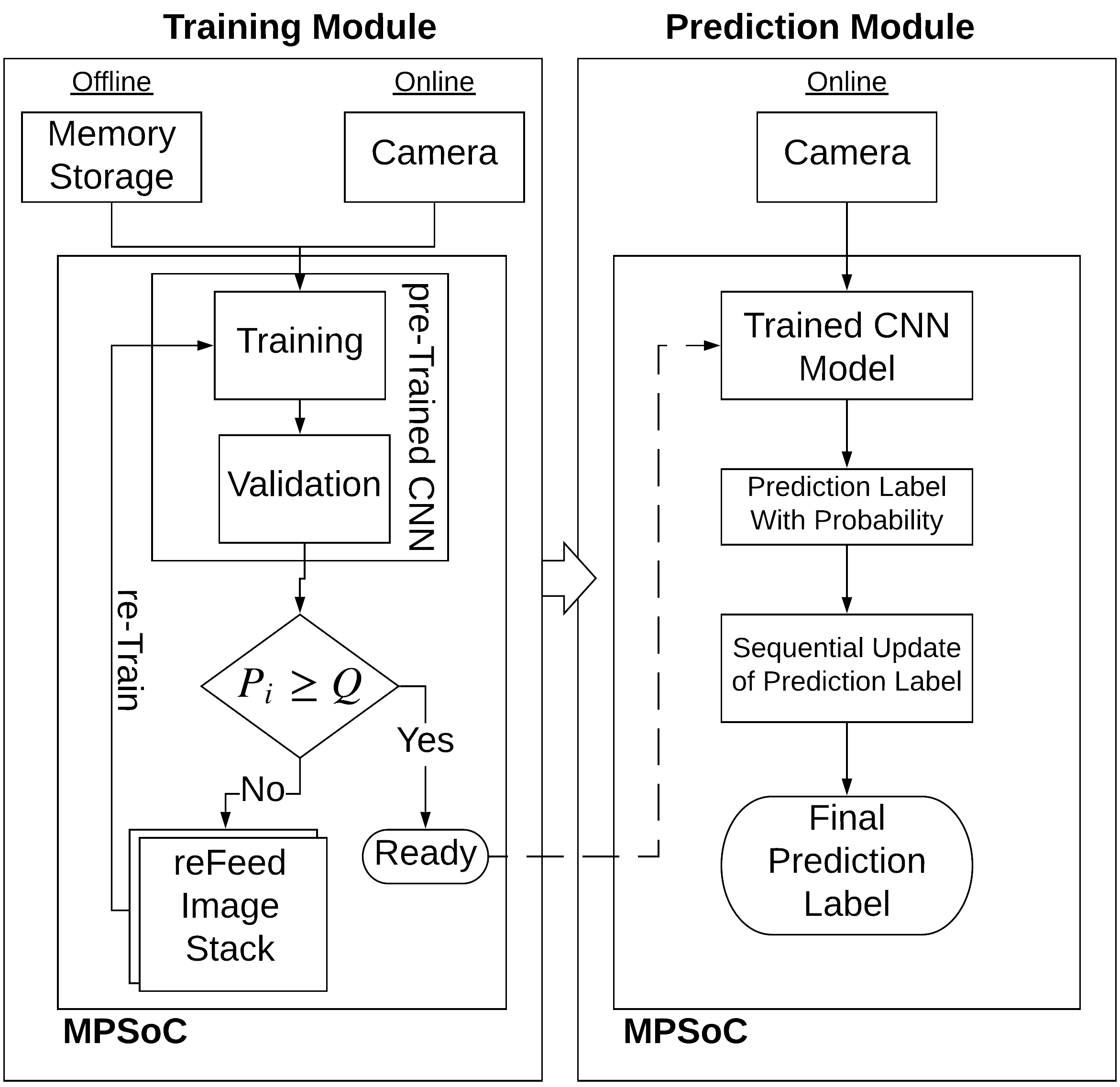}}
		\caption{IRON-MAN Model Work-flow}
		\label{cnnmpsoc}
	\end{figure}

	In our proposed approach, we utilize the concept of \textit{Hybrid Training Method} \cite{dey2018mat}, where we train our model both during offline (training period) and online (runtime/post-training period) modes. IRON-MAN has two modules in it: \textit{Training} and \textit{Prediction} (as shown in Fig. \ref{cnnmpsoc}). The strength of our approach is that it provides temporal analysis of videos without motion features and we call this type of prediction as \textbf{\textit{Temporal Motionless Analysis of Videos}} ($TMAV$).
	
	In the \textit{learning module}, we use \textit{transfer learning}\footnote{Learning achieved by taking the convolutional base of a pre-trained network, running the new data of 4 traffic categories through it and training a new randomly initialized classifier} \cite{transferLearning2010, zeiler2014transferLearning} by utilizing an existing pre-trained network and train the classifier with our data categories. First, we train the pre-trained CNN with our dataset, which could be either performed on the MPSoC or on a powerful computing system of which the computing resources could be leveraged to improve training time. After the initial phase of training is complete, we evaluate the overall prediction accuracy of the trained CNN. If the overall prediction accuracy ($P_i$) of the CNN is equal or more than the desired quality of experience ($Q$) \cite{dey2018mat}, then we utilize the CNN for prediction in the prediction module, but if the desired prediction accuracy is not achieved then we \textit{retrain} the CNN with the failed predicted images. This \textit{retrain} methodology is bio-inspired and mimics one of the key intelligence feature of a human being, which is learning from the surrounding environment to adapt. When a human being meets a new environment and is not aware of the rules and regulations associated with it, the human tries to adjust and adapt by learning the new set of rules and regulations by correcting its mistakes arising from the newly learnt inaccurate rules. We have utilized the same concept in our approach as well.
	
	Once, after the \textit{retraining} of the CNN, when the desired prediction accuracy is achieved we use the trained CNN in the \textit{prediction} module. Now, instead of providing prediction result for each individual image frame, we integrate the final prediction by taking previous image frames into account. Our CNN model's prediction is inspired by \textit{Bayes' theorem} and \textit{sequential Bayesian updating} \cite{doucet2000sequential}, where the model updates the possibility of the prediction label occurring by taking the probability of the label occurring in the previous frames. This approach is again bio-inspired and is adopted from the ideology of humans updating their knowledge using Bayesian logic. Detailed algorithm and inner working of the IRON-MAN is provided in the following two subsections.
	
	\subsection{Training Module}\label{para:train}
	For the proposed approach, any pre-trained CNN model such as VGG \cite{vgg2014}, ResNet \cite{resnet2016}, etc could be selected. The training module itself consist of two part training: Offline and Online mode. During the offline mode, the CNN is trained with stock images from a dataset stored on a memory. After the initial training period (offline) the CNN is then fed with live images from the camera and the prediction for each image frame is evaluated. Upon failure in prediction for each image, the image is stored in a stack implementation called ``\textit{reFeed Image Stack}". After utilizing cross-validation technique using live frames (images) from the camera, the overall prediction accuracy ($P_{i}$) of the CNN model is evaluated. If the overall prediction accuracy is equal or more than the desired quality of experience ($Q$) then the training process concludes and the CNN is ready to start predicting categories in the \textit{prediction} module. The governing equation to check for the suitability of the CNN for further prediction is provided in Eq. \ref{eqQoE}\cite{dey2018mat}, where $I$ is the dataset consisting of images, $i$ is any image in the dataset and $P_i$ is the prediction accuracy of the CNN for $i$. But if the desired quality of experience is not met in terms of prediction accuracy then the CNN is trained with the failed prediction images, which are stored in the \textit{reFeed image stack}. We call this as \textit{reTrain} approach so that the CNN can achieve a higher \textit{localized prediction accuracy}. Here, the term \textit{localized prediction accuracy} means the prediction accuracy of the model for a set of images that is restricted to a specific task or place. Algorithm for the Learning module is provided in Algo. \ref{algo1}.
	
	\begin{equation}\label{eqQoE}
	\forall \{i \in I : i > 1\}, P_{i} \geq Q
	\end{equation}
	
	\noindent \textbf{Note}: Using the proposed approach, training could be performed both on the MPSoC or on a more powerful computing device. If the CNN is trained on a device other than the MPSoC, then after the aforementioned training phases (offline and online) are complete, the CNN model's parameters and weights could be saved and migrated to the MPSoC to act as the \textit{Prediction} module. Utilizing this method of training the CNN can also improve reliability (lifespan) and energy efficiency of the device due to the required operating resources during the training period being high. Here, the operating resources represent the computing resources such as CPU, GPU, memory, etc. required during the operation of an executing application.
	
	\begin{algorithm} [!h]
		\small{
			\KwIn{\\1. $\mathcal{I}$: set of $n$ Images from training \& validation dataset\\
				2. $\mathcal{C}$: set of $m$ Images from camera (for cross-validation)}
			\KwOut{$P$: prediction accuracy after training }
			
			\textbf{Initialize:} $Q$ = 0.7;     \Comment{Quality of experience is set to 70\% by default}\\
			$\mathcal{S}.Count$ = 0; \Comment{$\mathcal{S}$: \textit{reFeed image stack}}\\
			\smallskip
			\underline{\textbf{Offline Training:}}\\
			Train~(pre-trained CNN model~, ~$\mathcal{I}$); \Comment{Train model with $\mathcal{I}$ dataset}\\
			\smallskip
			\underline{\textbf{Online Training:}}\\
			\For{each image $i \in \mathcal{C}$}
			{Prediction = Test~(~CNN model~); \Comment{Test(CNN model) is a function which outputs whether prediction is correct or wrong}\\
				\Comment{Prediction.IsWrong() is a function to return True when Prediction.Label $!=$ Original.Label of test image $i$}\\
				\If{Prediction.IsWrong()}{$S.Push(i)$;}
				$P_{i}$ = CalculateMeanPredictionAccuracy();\\ \Comment{CalculateMeanPredictionAccuracy() is the function to evaluate the overall prediction accuracy of the CNN and return the value ($P_{i}$)}\\
			}
			\{\textit{re-Train with reFeed Image Stack if $P_{i} < Q$}\}\\
			\If{$P_{i} < Q$}{\Comment{Need to satisfy condition of Eq. \ref{eqQoE}}\\
				\If{S.Empty() == False}{
					\{\textit{Traing CNN with \textit{reFeed image stack}}\}\\
					Train~(CNN model~,~~$\mathcal{S}$ );\\
					$P_{i}^{f}$ = CalculateMeanPredictionAccuracy();\\ \Comment{CalculateMeanPredictionAccuracy() is the function to evaluate the overall prediction accuracy of the CNN after retraining and return the value ($P_{i}^{f}$)}\\
				}
				$P = P_{i}^{f}$;\\
				$\mathcal{S}.Count$ = 0; \Comment{reset \textit{reFeed image stack}}
			}
			\Else{return $P$;}
			
			\caption{Training Module Execution}    
			\label{algo1}
		}
	\end{algorithm}

	\subsection{Prediction Module}
	From Bayes' Theorem \cite{doucet2000sequential} we can represent the expression representing the \textit{Bayesian Update Scheme} is as follows:
	
	\begin{equation}\label{bayeseq}
	posterior \propto prior \times likelihood
	\end{equation}
	
	\noindent In Eq. \ref{bayeseq}, \textit{posterior} is the revised probability of an event occurring after taking new information into consideration, \textit{prior} is the probability of the event assessed before revising posterior and \textit{likelihood} is the probability that an event which has already occurred would yield a specific outcome. Now, using sequential update scheme where we take the past into account and the modified expression for Bayesian Update Scheme is as follows:
	
	\begin{equation}\label{newbayes}
	new~posterior \propto current \times new~likelihood
	\end{equation}
	
	In Eq. \ref{newbayes}, \textit{current} is the probability of some entity occurring whereas the \textit{new likelihood} is the Bayesian Update taking posterior from the past into account. This approach sometimes is also called a Recursive Bayesian Update. For our scenario, we are trying to predict the current probability for the label (category), which becomes the \textit{new posterior} in the equation, \textit{current} is the probability prediction of the category of the image frame provided by the CNN and \textit{new likelihood} is the probability of the category occurring in some previous time steps. Here, the reason to mention \textit{some previous time steps} is because the number of previous time steps to take into consideration will be a heuristic choice of the user. In our case, we call the number of previous frames (images), which is considered to provide an integrated prediction for the chosen category, as \textit{Frame Window}. \textit{Frame Window} consists of \textit{N} number of frames, which are taken into consideration.
	
	If we consider that the prediction for the category in the current frame as $P_{this}^{category}$, prediction for the same category in the previous frame as $P_{this - 1}^{category}$ and the total prediction accuracy of the model as $P_{CNN}$ then the updated equation for Bayes' Theorem is as follows:
	
	\begin{equation}\label{bayestheorem}
	P_{updated}^{category} \propto P_{this-1}^{category} \times P_{this}^{category}
	\end{equation}
	
	Eq. \ref{bayestheorem} could be utilized to predict the frame using the prediction of previous frame as follows:
	
	\begin{equation}\label{bayesmaineq}
	P_{updated}^{category} = \frac{P_{this-1}^{category} \times P_{this}^{category} }{ (P_{this-1}^{category} \times P_{this}^{category}) + P_{CNN}}
	\end{equation}
	
	In Eq. \ref{bayesmaineq}, $P_{updated}^{category}$ is the updated prediction using Bayes' Theorem for the same category by the CNN model. We should also note that both $P_{this-1}^{category}$ and $P_{this}^{category}$ are \textit{conjugate priors} for our scenario since they belong to the same category as the posterior ($P_{updated}^{category}$) and hence in the same probability distribution family. Now, depending on the \textit{Frame Window}, the evaluation of $P_{updated}^{category}$ will vary, which leads us to an updated equation as follows:
	
	
	\begin{equation}\label{maineq}
	P_{updated}^{category} =
	\begin{cases}
	\vspace{4mm}
	P_{this}^{category}, & \text{if}\ N = 0 \\ \vspace{4mm}
	\frac{P_{this-1}^{category} \times P_{this}^{category} }{ (P_{this-1}^{category} \times P_{this}^{category}) + P_{CNN}}, & \text{{if}\ N = 1}\\ \vspace{4mm}
	\prod_{1}^{N}\frac{P_{this-N}^{category} \times P_{this-(N-1)}^{category} }{ (P_{this-N}^{category} \times P_{this-(N-1)}^{category}) + P_{CNN}}, & \text{if}\ N > 1 \\
	\end{cases}
	\end{equation}
	
	Eq. \ref{maineq} is the governing equation, which is utilized to predict the probability of the category during the \textit{Frame Window}. 
	
	In the Prediction module, IRON-MAN has a queue implementation of the image stack (called as \textit{Image Queue}), where the $N$ number of frames are stored and $N$ is defined by the user to denote the size of the \textit{Frame Window}. When an ${(N+1)}^{th}$ image frame comes from the camera for prediction, the images stored at $1^{st}$ position of the \textit{Image Queue} is popped out and the ${(N+1)}^{th}$ image frame is pushed in the $N^{th}$ position of the queue while everything getting shifted a place in the middle just like in a first-in-first-out (FIFO) queue implementation. When prediction for a particular frame is required, the prediction of the frame by the CNN model is provided as well as the prediction of the \textit{Frame Window} is provided by using the Eq. \ref{maineq}. After utilizing Eq. \ref{maineq} the updated prediction ($P_{updated}^{category_i}$) for a specific category $i$ is compared with the the updated prediction of other categories and the label for the maximum value of the prediction is provided as output. The detailed algorithm for the Prediction module is provided in Algo. \ref{algo2}.
	
	\begin{algorithm} [!h]
		\small{
			\KwIn{\\$\mathcal{N}$: number of frames in the \textit{frame window}}
			\KwOut{\\ 1. $P_{label}$: prediction label \\
				2. $P_{updated}^{category}$: updated prediction label using Eq. \ref{maineq}
			}
			\smallskip
			\textbf{Initialize:}
			$\mathcal{I}.Count$ = 0; \Comment{$\mathcal{I}$: \textit{Image queue} of size \textit{N}}\\
			\smallskip
			\While{Camera is feeding image frames}
			{
				\If{$\mathcal{I}.Count > $ 0}
				{
					$\mathcal{I}$.Pop();\\ \Comment{Pop() is a function to remove the object/image in the 1st position of the \textit{image queue}}
				}
				$\mathcal{I}$.Push(Current Frame);\\ \Comment{Push(Image) is a function to remove the object/image in the $N^{th}$ position of the \textit{Image Queue}}\\
				\vspace{2mm}
				$P_{label}$ = PredictCurrentFrame();\\ \Comment{PredictCurrentFrame() is a function to predict the label (category) for the current frame}\\
				\vspace{2mm}
				$P_{updated}^{category}$ = UpdatedPrediction($N$);\\ \Comment{UpdatedPrediction($N$) is a function to predict the label taking prior $N$ sequential frames into account by utlizing the Eq. \ref{maineq} for all available categories and retunring the label for the category with higgest probability}\\
			}
			
			\caption{Prediction Module Execution}    
			\label{algo2}
		}
	\end{algorithm}
	
	\subsection{ECTI: A new metric to choose suitability of a CNN model in embedded systems}
	If we consider $ET$ as the total execution time period required to train the CNN with a dataset $I$ consisting of $n$ number of images to achieve a validation prediction accuracy of $P$, $Q$ as the \textit{quality of experience}, and the average power consumption per second during the training period as $e$ then the equation for \textit{Energy Consumption per Training Image} ($ECTI$) could be defined as follows:
	
	\begin{equation}\label{ectieq}
	ECTI = (\frac{ET}{n} \times e) ~\text{iff}~P \ge Q
	\end{equation}
	
	The unit of $ECTI$ is $kilo-watt-hour$ ($kWh$), where $ET$ is represented in hours and $e$ in \textit{kilo-Watt} ($kW$). To choose the most suitable CNN for an embedded application we have to select the CNN with the least value of $ECTI$.
	
	\begin{table*}[!h]
		\centering
		\begin{tabular}{ | c | m{6cm} | m{6cm} | }
			\hline
			Image frames & \centering Normal predictions without IRON-MAN & Predictions by IRON-MAN \\ \hline
			\begin{minipage}{.21\textwidth}
				\centering
				\includegraphics[width=3cm,height=2.3cm]{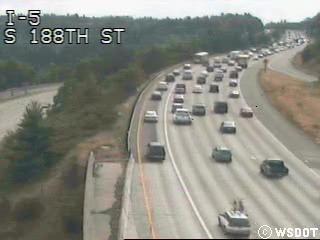}
			\end{minipage}
			&
			\centering
			\underline{\textbf{(Label) $\bigm\lvert$\hspace{1mm} Empty $\bigm\lvert$\hspace{1mm} Fluid $\bigm\lvert$\hspace{1mm} Heavy $\bigm\lvert$\hspace{1mm} Jam}}\\
			\smallskip \smallskip \smallskip
			(Fluid) | 0.0100 | \textbf{0.7930} | 0.1683 | 0.0286
			& 
			\underline{\textbf{(Label) $\bigm\lvert$\hspace{1mm} Empty $\bigm\lvert$\hspace{1mm} Fluid $\bigm\lvert$\hspace{1mm} Heavy $\bigm\lvert$\hspace{1mm} Jam}} \newline
			\newline
			(Fluid) $\bigm\lvert$ 0.01 $\bigm\lvert$ \textbf{0.793} $\bigm\lvert$ 0.1683 $\bigm\lvert$ 0.0286
			\\
			\begin{minipage}{.21\textwidth}
				\centering
				\includegraphics[width=3cm,height=2.3cm]{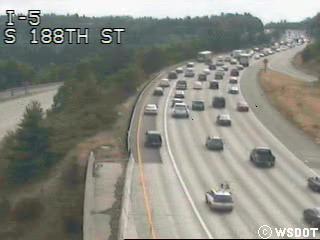}
			\end{minipage}
			&
			\centering
			(Heavy) $\bigm\lvert$ 0.0131$\bigm\lvert$ 0.3091$\bigm\lvert$ \textbf{0.5098}$\bigm\lvert$ 0.1681
			& 
			(Fluid) $\bigm\lvert$ 0.0001 $\bigm\lvert$ \textbf{0.1986} $\bigm\lvert$ 0.0798 $\bigm\lvert$ 0.0048
			\\ 
			\begin{minipage}{.21\textwidth}
				\centering
				\includegraphics[width=3cm,height=2.3cm]{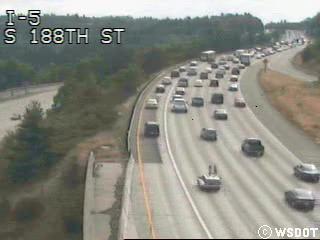}
			\end{minipage}
			&
			\centering
			(Jam) $\bigm\lvert$ 0.0131 $\bigm\lvert$ 0.2754 $\bigm\lvert$ 0.3399 $\bigm\lvert$ \textbf{0.3717}
			& 
			(Fluid) $\bigm\lvert$ 0.0 $\bigm\lvert$ \textbf{0.0524} $\bigm\lvert$ 0.0267 $\bigm\lvert$ 0.0018
			\\ \hline
		\end{tabular}
		\caption{Predictions of traffic categorization image frames with and without IRON-MAN}\label{tbl:result1}
	\end{table*}
	
	\begin{table*}[h!]
		\centering
		\begin{tabular}{ | c | m{6cm} | m{6cm} | }
			\hline
			Image frames & \centering Normal predictions without IRON-MAN & Predictions by IRON-MAN \\ \hline
			\begin{minipage}{.21\textwidth}
				\centering
				\includegraphics[width=3.4cm,height=2.3cm]{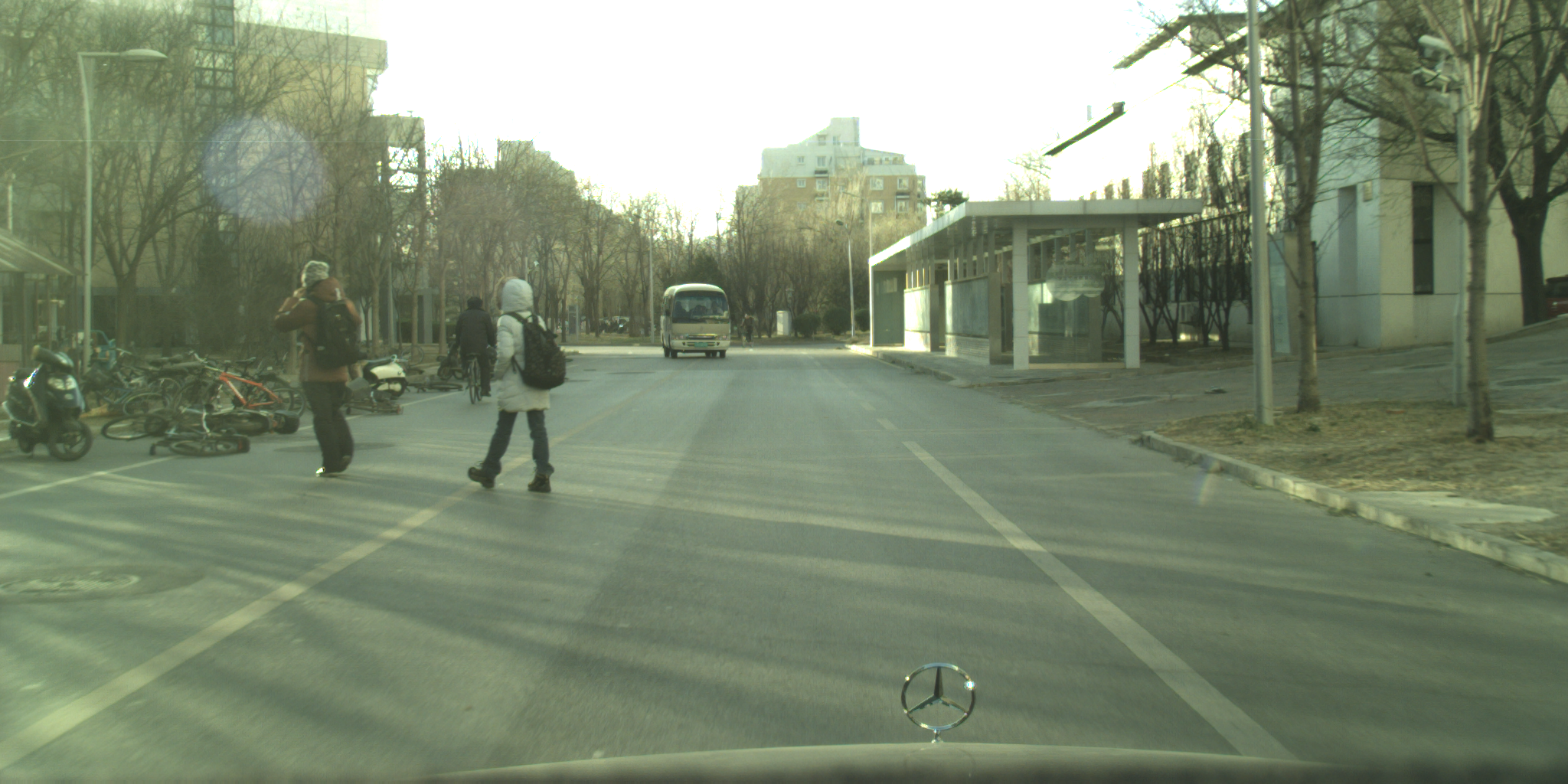}
			\end{minipage}
			&
			\centering
			\underline{\textbf{(Label) $\bigm\lvert$\hspace{2mm} No-Obstruction $\bigm\lvert$\hspace{1mm} Obstruction }}\\
			\smallskip \smallskip \smallskip
			(Obstruction) $\bigm\lvert$\hspace{4mm} 0.3270 $\bigm\lvert$\hspace{6mm} \textbf{0.6730} 
			& 
			\underline{\textbf{(Label) $\bigm\lvert$\hspace{2mm} No-Obstruction $\bigm\lvert$\hspace{1mm} Obstruction }} \newline
			\newline
			(Obstruction) $\bigm\lvert$ \hspace{3mm} 0.3270 $\bigm\lvert$ \hspace{12mm} \textbf{0.6730} 
			\\
			\begin{minipage}{.21\textwidth}
				\centering
				\includegraphics[width=3.4cm,height=2.3cm]{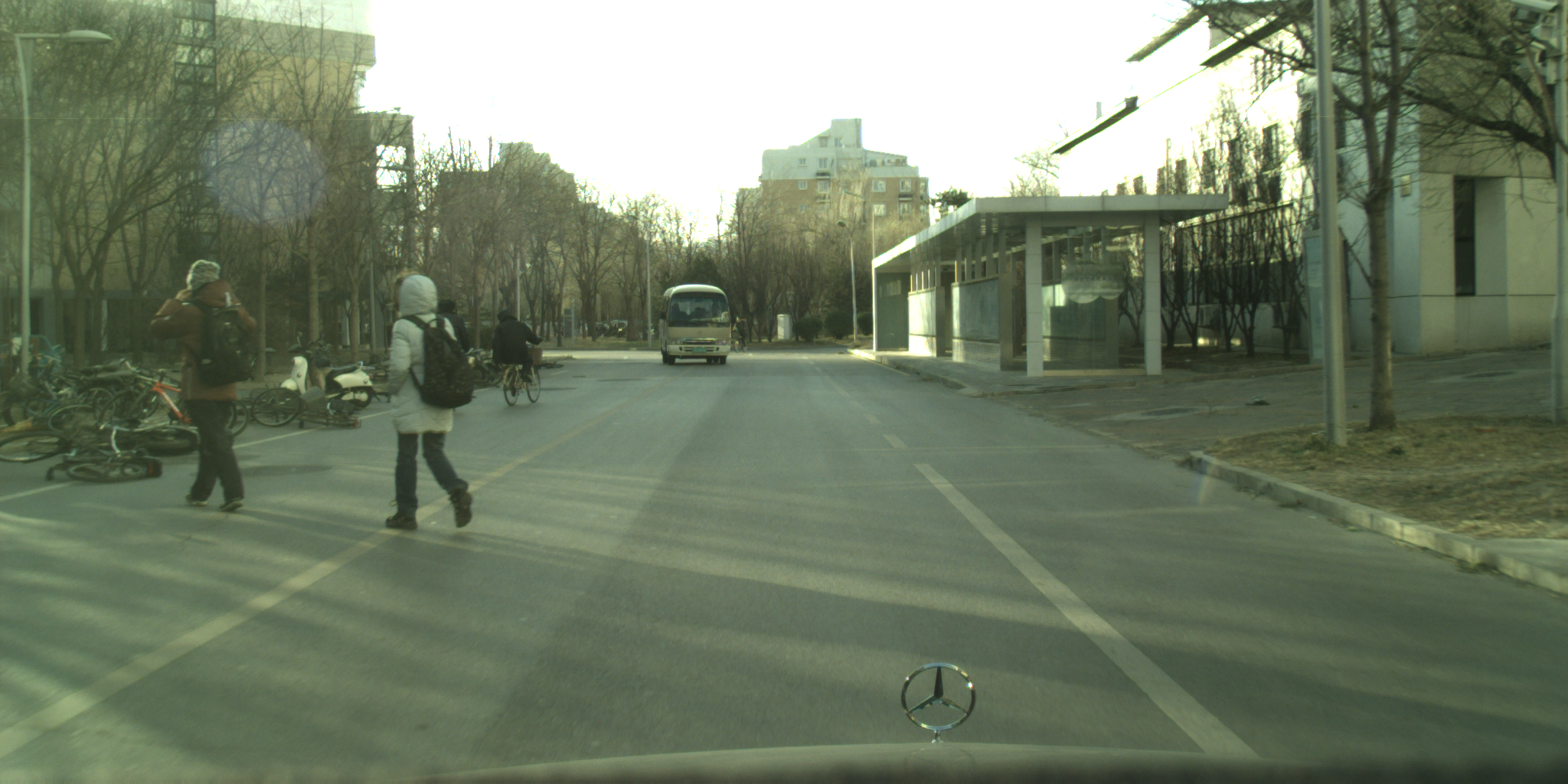}
			\end{minipage}
			&
			(No-Obstruction) $\bigm\lvert$ \textbf{0.5010} \hspace{1mm}$\bigm\lvert$\hspace{4mm} 0.4990 
			& 
			(Obstruction) $\bigm\lvert$\hspace{4mm} 0.1843 $\bigm\lvert$\hspace{12mm} \textbf{0.3166} 
			\\ 
			\begin{minipage}{.21\textwidth}
				\centering
				\includegraphics[width=3.4cm,height=2.3cm]{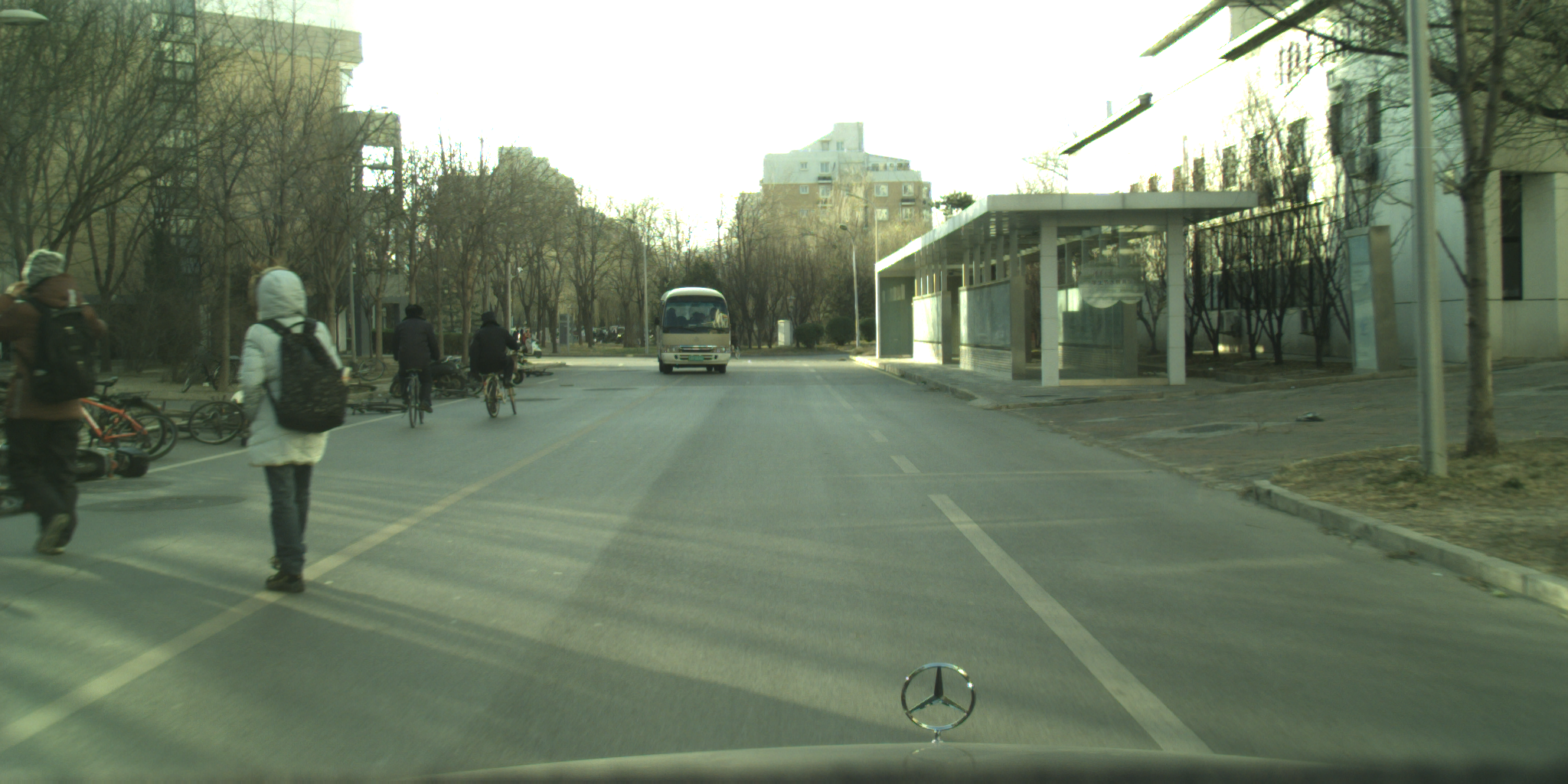}
			\end{minipage}
			&
			\centering
			(Obstruction) $\bigm\lvert$\hspace{4mm} 0.4600 $\bigm\lvert$\hspace{3mm} \textbf{0.5400}
			& 
			(Obstruction) $\bigm\lvert$\hspace{4mm} 0.1047 $\bigm\lvert$\hspace{12mm} \textbf{0.1908} 
			\\ \hline
		\end{tabular}
		\caption{Predictions of pedestrian obstruction image frames with and without IRON-MAN}\label{tbl:result2}
	\end{table*}
	
	\section{Experimental Results}\label{results}
	
	\subsection{Dataset Used}\label{dataset}
	We have performed our validation on two different test cases: \textit{traffic categorization} and \textit{pedestrian obstruction}.
	
	\subsubsection{Traffic categorization dataset}
	For our \textit{traffic categorization} experimentation we are using the same dataset used by Luo et al. \cite{luo2015traffic, luo2016traffic} and Dey et al. \cite{dey2018mat} for traffic categorization to validate the performance of our proposed methodology. Mainly two dataset are used in this experiment. The first dataset is released by UCSD traffic control department \cite{datasetChan2005classification}. This dataset contains 254 highway video sequences, all of which are filmed using the same camera containing light, heavy and traffic jams filmed at different periods of the day under different weather conditions. Each UCSD video has a resolution of 320 $\times$ 240 pixels with a frame rate of 10 fps. The video streams from the UCSD dataset were converted  to images by processing 1 frame out of every 8 frames (\textasciitilde1.3 fps). This UCSD dataset was used as the testing dataset. Since UCSD dataset is categorized into 3 labels: \textit{Light, Medium and Heavy}, we manually annotated the images into our desired 4 categories: \textit{Jam, Heavy, Fluid, Empty}. Another dataset consisting of the 400 images, which is provided by Luo et al.\cite{luo2015traffic} as well as Dey et al.\cite{dey2018mat}, captured from highway cameras deployed all over the UK and also consist of several examples of different weather and lighting conditions in order to provide a better training performance. These 400 images are segregated into 4 categories: \textit{Jam, Heavy, Fluid, Empty}; and each category having 100 images. This dataset was used for training and validation purposes. When the road was empty the images were categorized as Empty and for light traffic it is labeled as Fluid. When the traffic was heavy the images are labeled as Heavy and for slow moving heavy traffic the images are labeled as Jam.
	
	\subsubsection{Pedestrian obstruction dataset}
	For our \textit{pedestrian obstruction} experimentation we have used the pedestrian dataset used by Li et al. \cite{li2016new}, which consists of more than 5 million images for a total of 32361 labeled vulnerable road users (VRUs), including cyclists, pedestrians, tri-cyclists and motor-cyclists etc. To validate our proposed IRON-MAN approach we have only used the training dataset of \cite{li2016new} consisting of 9742 images for both training and validation, whereas we used the validation dataset from the same study for testing purposes. The images were manually labeled into 2 categories: \textit{Obstruction, No-Obstruction}. Whenever there were pedestrians in front or nearby the camera view, then it is categorized as obstruction and in contrary it is labeled as no-obstruction.
	
	\subsection{Hardware Setup}
	We have implemented the methodology on an Odroid-XU4 \cite{odroidxu4}, which employs Exynos 5422 MPSoC\cite{exynos5422} used in popular Samsung Note phones and phablets. Exynos 5422 implements ARM's big.LITTLE architecture utilizing 4 ARM Cortex A-15 big CPUs, 4 ARM Cortex A-7 LITTLE CPUs and 6 MALI T628 MP6 GPUs. The Odroid-XU4 does not have an internal power sensor, and we had to use an external power monitor \cite{odroidsmartpower2} with networking capabilities over WIFI to take power consumption readings. Odroid-XU4 platform has 4 temperature sensors on 4 ARM Cortex A-15 big CPU cores, which we have read to monitor temperature behavior during our experiments. For all our experiments we have only utilized 4 A-15 big CPU cores to monitor the operating temperature and power consumption.
	
	
	\subsection{Experimental Results}\label{resultSection}
	
	\subsubsection{Prediction accuracy evaluation}\label{prediction_accuracy}
	We chose two pre-Trained CNN models, which were trained on millions of ImageNet images, for our validation. These two CNN models are VGG16 \cite{vgg2014} and ResNet50 \cite{resnet2016}. In our experiments, we have chosen the $quality~of~experience$ ($Q$) to be 0.7 i.e. 70\%. For VGG16 it took us 360 images to train the pre-Trained CNN for traffic categorization using Transfer Learning and retrain approaches, (see Sec. \ref{para:train}) and gained a testing prediction accuracy of 98.93\%. The \textbf{average power consumption was 10.63 W} on the Odroid-XU4 MPSoC and the total execution time was 97 minutes 44 seconds. Therefore, using the formula in Eq. \ref{ectieq}, the \textit{Energy Consumption per Training Image} comes out to be \textbf{$48.097 \times 10^{-6}~kWh$} (average) ($\simeq (10.63 \times \frac{1}{1000}) \times \frac{(97 \times 60 + 44)}{3600} \times \frac{1}{360}$). 
	
\begin{figure}
		\centering
		\includegraphics[width=9cm]{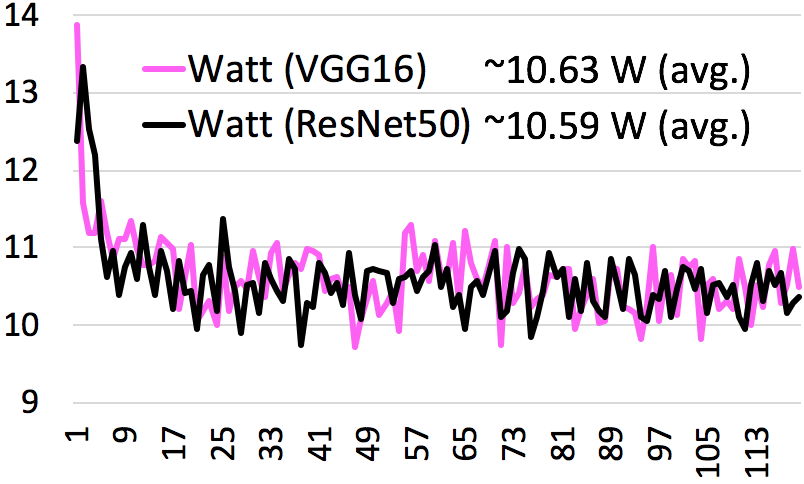}
		\caption{Power consumption for VGG16 and ResNet50 in Watts (Watt vs time interval)}
		\label{powergraph}
	\end{figure}

	For ResNet50 it took us 117 minutes and 27 seconds, and 330 images to train the pre-Trained CNN for traffic categorization using transfer learning and retrain approaches. The testing prediction accuracy attained by the model is 92.79\%. The \textbf{average power consumption was 10.59 W}. Therefore, using Eq. \ref{ectieq}, the \textit{Energy Consumption per Training Image} comes out to be \textbf{$62.817 \times 10^{-6}~kWh$} (average) ($\simeq (10.59 \times \frac{1}{1000}) \times \frac{(117 \times 60 + 27)}{3600} \times \frac{1}{330}$). Since both VGG16 and ResNet50 were able to achieve a prediction accuracy higher than $Q$ (70\%) and the ECTI of VGG16 is lower than ResNet50, we have chosen VGG16 as a more suitable CNN model for training on the embedded device, i.e. Odroid-XU4. During training VGG16 and ResNet50, the average maximum temperature achieved by the big CPU cores are 93.60\degree C (average) and 93.72\degree C (average) respectively. The maximum operating temperature of the big CPU cores for different CNN models are shown graphically in Fig. \ref{tempgraph}. We noticed that the maximum \textit{baseline temperature}, which is the operating temperature of the CPU core when idle i.e. only executing background tasks while on Linux's ondemand power scheme, of the ARM Cortex A-15 big CPU core was 69.24\degree C (average). Therefore, the deviation of operating temperature while training and the baseline temperature was 24.36 \degree C (average) for VGG16 and 24.48 \degree C (average) for ResNet50.  Max T. and Max Baseline T. stands for maximum operating temperature and maximum baseline temperature in Fig. \ref{tempgraph}. In Fig. \ref{powergraph} we show graphical representation of power consumption for both VGG16 and ResNet50 training period. For both Fig. \ref{tempgraph} and Fig. \ref{powergraph} we have only shown a snapshot of the total training execution period due to repetition in behavior of the graph.
	
	Now, to prove efficacy of our integrated rational prediction of image frames from the video, we randomly chose a video from UCSD traffic dataset\cite{datasetChan2005classification} and broke the video into image frames, which represents the same category as the video itself. We chose a video from medium traffic category, which corresponds to fluid category and utilized the VGG16 model as the CNN model in IRON-MAN to predict the label of a sequence of image frames as well as the label for video. For the experiment we chose the \textit{Frame Window} of 3 images representing a video sequence of 4 seconds (approx.). Table. \ref{tbl:result1} shows the prediction of labels if we only use VGG16 CNN without our IRON-MAN approach and if we use IRON-MAN as well. The table shows that VGG16 could predict the label of the image frame, but that prediction did not match the label for the whole video, whereas the prediction label for the video using IRON-MAN was achieved with 100\% accuracy.
	
	In another experiment (\textit{pedestrian obstruction}), we chose simultaneous image frames having pedestrians as obstruction from the pedestrian dataset\cite{li2016new} to validate efficacy of IRON-MAN. Table \ref{tbl:result2} shows that IRON-MAN was again able to predict whether the path is obstructed by a pedestrian or
	not using integrated results from previous frames. ResNet50 CNN was utilized in this experiment, which gained a testing prediction accuracy of 72.50\% after \textit{transfer learning} and \textit{retraining} approaches for this application. For pedestrian obstruction experiment we trained the classifier of ResNet50 and VGG16 on a general purpose computer instead due to massive training dataset and associated energy consumption, and migrated the saved parameters and weights of the model to Odroid-XU4 for further prediction. ResNet50 was selected over VGG16 in this experiment because the model gained a better testing prediction accuracy for this application, which outperformed VGG16's 66.87\% testing prediction accuracy for the same.
	
	\subsubsection{Comparison between traditional method \& IRON-MAN}
	If we consider a video consisting of $M$ number of image frames need to be analyzed and predicted using CNN based approaches such as VGG and ResNet, traditional method such as VGG or ResNet would analyze each individual image frames from $M$ number of images and output the result. Whereas, from the results of experiments provided in Sec. \ref{prediction_accuracy}, we could notice that \textit{IRON-MAN} is capable of analyzing $N$, where $N \le M$, number of image frames together and provides a consolidated prediction of $N$ frames. Therefore, \textit{IRON-MAN} is able to categorize video with 100\% prediction accuracy during testing and is able to perform \textit{TMAV}, which could not be performed by traditional CNN based approaches.
	
	\subsubsection{Comparative study of IRON-MAN}
	To evaluate the efficacy of \textit{IRON-MAN}, we compared the methodology with the state-of-the-art approach for traffic categorization proposed by Luo et al. \cite{luo2018traffic}. Since, the methodology proposed in \cite{luo2018traffic} is closely related to the target application of traffic categorzation without motion features, it is justifiable to use the proposed methodology  for a comparative study with \textit{IRON-MAN}. Additionally, \textit{IRON-MAN} being the first methodology to perform \textit{TMAV} in traffic categorization, it would be unjustified to compare the methodology with any other existing approaches. 
	
	In \cite{luo2018traffic}, the researchers have used SegCNN and RegCNN to analyze and categorize traffic. The main motivation of utilizing SegCNN and RegCNN in their approach is to improve the accuracy of the prediction model even without training the CNN with a large dataset.
	
	Although after the training, Luo et al.'s approach was able to achieve a prediction accuracy of 94.8\% during testing for traffic categorization, which is slightly higher ($\simeq2.01$\%) than the prediction accuracy achieved by \textit{IRON-MAN}, but when the approach was tested for temporal analysis of image frames to provide holistic prediction of the video instead, it achieved a prediction accuracy of 65\% for the same target application. In comparison \textit{IRON-MAN} was able to achieve 100\% prediction accuracy for temporal analysis of 9 sequential image frames in the \textit{frame window} as mentioned in Sec. \ref{resultSection}. 
	
\begin{figure}
		\centering
		\includegraphics[width=7.9cm]{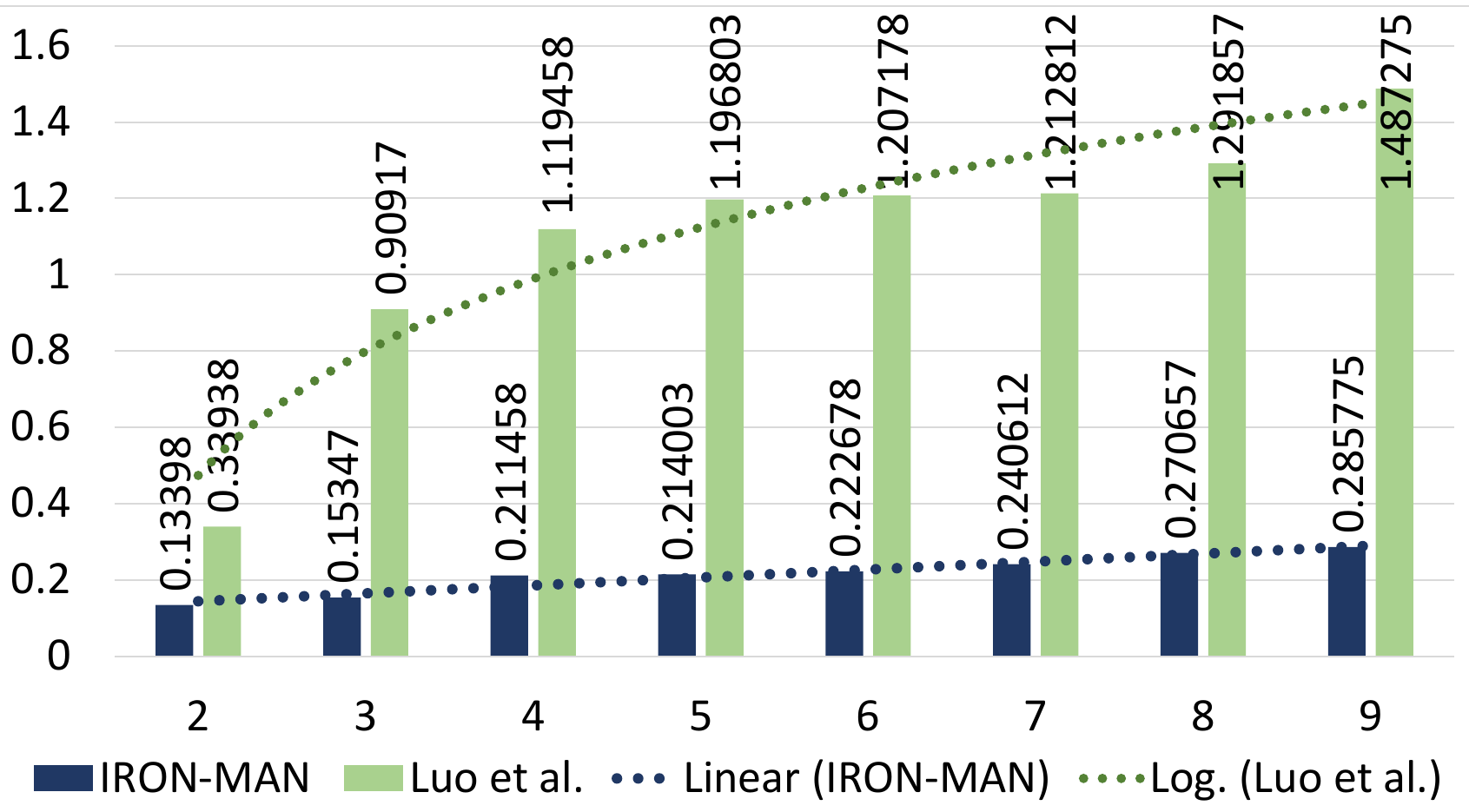}
		\caption{Execution time taken to analyze 9 image frames sequentially from the traffic video (Execution time in seconds vs number of image frames analyzed)}
		\label{comparison_graph}
	\end{figure}

	Moreover, given the computation complexity of \cite{luo2018traffic}, executing the approach also has an overhead of 5.2$\times$ on the Odroid XU4 MPSoC than compared to \textit{IRON-MAN}. Fig. \ref{comparison_graph} shows the execution time (in seconds) of analyzing 9 sequential image frames of the same video from the UCSD dataset \cite{datasetChan2005classification}. Each test was performed 5 times on the Odroid XU4, utlizing all eight CPU cores (big.LITTLE CPU cores) and the average execution time for image analysis for each number of image frame is provided in Fig. \ref{comparison_graph}. Since the concept of \textit{frame window} or temporal analysis of image frames of video does not exist in \cite{luo2018traffic} approach, we sequentially feed each image frames from the video to the methodology to get the output analysis.
	
	From Fig. \ref{comparison_graph} we can notice that it takes 1.487275 seconds for Luo et al.'s approach to analyze 9 sequential image frames, whereas it takes 0.285775 seconds to analyze 9 sequential image frames in the \textit{frame window} for the \textit{IRON-MAN}. Therefore, the overhead associated with Luo et al.'s approach is 5.2$\times$ (approx.) on the Odroid XU4 MPSoC, making the \textit{IRON-MAN} more suitable for real-time execution for traffic categorization application on the embedded devices utilizing MPSoC.
	
		\begin{figure}
		\centering
		\includegraphics[width=9cm]{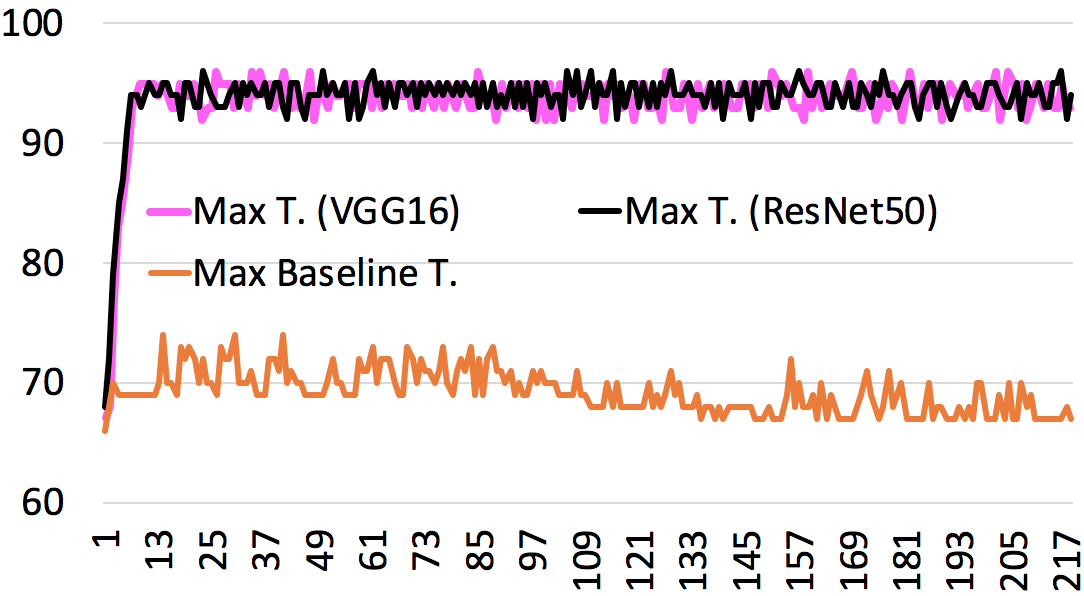}
		\caption{Graphical representation of operating temperature peaks of VGG16 and ResNet50 training and baseline temperature (\degree C vs time interval)}
		\label{tempgraph}
	\end{figure} 
	
	\subsubsection{Effect on lifespan of the device}
	In the studies \cite{chantem2008temperature, coskun2007temperature, zhou2016thermal}, it has been shown that an increase in the operating temperature of an embedded device by 10-15\degree C could reduce the lifespan of the device by 2$\times$. In Sec. \ref{resultSection} and from the knowledge on thermal deviation in Fig. \ref{tempgraph}, we have already noticed that if training is performed on the device then there is an increase in operating temperature of the device by 24.36\degree C (average) for VGG16 and 24.48\degree C (average) for ResNet50. Therefore, considering that the device lifespan reduces by 2$\times$ for every 10\degree C increase in operating temperature and if training is performed on the device then the lifespan of the same device reduces by 4.872$\times$ ($\simeq \frac{24.36}{10} \times 2$) for utilizing VGG16 and 4.896$\times$ ($\simeq \frac{24.48}{10} \times 2$) for utilizing ResNet50. Now, if we consider that lifespan reduces by 2$\times$ for an increase of 15\degree C in operating temperature instead, then the same evaluation changes to 3.248$\times$ for VGG16 and 3.264$\times$ for ResNet50. Therefore, until the application requires the CNN to be trained on the embedded MPSoC to continue providing desirable analysis it is highly recommended that the CNN is not trained on the embedded system.
	
\section{Limitations and Discussion}
\begin{lemma}\label{lemma1}
	Updated prediction ($P_{updated}^{category}$) of the category using Baye's Theorem tends to zero as prediction ($P_{this}^{category}$) for the category in the current frame tends to zero.
\end{lemma}

In the Eq. \ref{bayestheorem}, if $P_{this}^{category} \longrightarrow 0$ then $P_{updated}^{category} \longrightarrow 0$ (see Eq. \ref{eqTendsToZero}). Hence, for a lower value of $P_{this}^{category} $ we would be achieving a lower value of $P_{this}^{category}$ if prediction ($P_{this-1}^{category}$) for the same category in the previous image frame is less than or equal to one\footnote{Prediction of an image frame could not be more than one since one represents 100\% probability of the category occurring and probability could only range from 0 to 100.} ($P_{this-1}^{category} \le 1$).

\begin{equation}\label{eqTendsToZero}
P_{updated}^{category}  \longrightarrow 0 ~\text{as}~ P_{this}^{category} \longrightarrow 0
\end{equation}

\begin{theorem}
	If $N \ge 1$ in the \textit{frame window} and $P_{this}^{category} \longrightarrow 0$ then the updated prediction ($P_{updated}^{category}$) of the category converging to zero increases as $N$ increases.
\end{theorem}

In Eq. \ref{maineq}, if the number of image frames ($N$) is greater than one then the updated prediction for the category is represented by Eq. \ref{eqMainGeN}.

\begin{equation}\label{eqMainGeN}
P_{updated}^{category} = \displaystyle \prod_{1}^{N}\frac{P_{this-N}^{category} \times P_{this-(N-1)}^{category} }{ (P_{this-N}^{category} \times P_{this-(N-1)}^{category}) + P_{CNN}},  \text{if}\ N > 1 \\
\end{equation}

In the Eq. \ref{eqMainGeN}, if we consider the term, $\frac{P_{this-N}^{category} \times P_{this-(N-1)}^{category} }{ (P_{this-N}^{category} \times P_{this-(N-1)}^{category}) + P_{CNN}}$ as $P_k$, since the preceding term is part of the product series represented in Eq. \ref{eqMainGeN} and consider $P_{updated}^{category}$ as a function ($F_{N,i}$) of $N$ and $i$, where $i$ is the $i_{th}$ image in the frame window consisting of N image frames, and both $i$ and $N$ are whole numbers ($N, i \in W$ and $N > i \ge 1$), then the aforementioned equation (Eq. \ref{eqMainGeN}) could be represented as follows:

\begin{equation}\label{eqModMainGen}
F_{N, i} = \displaystyle \prod_{k = N - i + 1}^{N} P_k
\end{equation}

Since $P^{category}_{this}$ is an integral part of $P_k$ in Eq. \ref{eqModMainGen} (see Eq. \ref{eqMainGeN}), using the knowledge from \textit{lemma 1} we can say that as $P^{category}_{this} \longrightarrow 0$ then $P_k \longrightarrow 0$ as well. Now, as $N \longrightarrow \infty$ and $P_k \longrightarrow 0$ then $F_{N, i} \longrightarrow 0$. Therefore, using this knowledge we could state that: $0 \le F_{N, i} \le P_k$ and proving the fact that as $N$ gets larger and $P^{category}_{this}$ gets smaller (close to zero), the updated prediction ($P^{category}_{updated}$) also converges to zero.
	
Through our empirical data we noticed that the aforementioned theorem holds true and this could be considered as a potential limitation of using Bayes' Theorem along with CNN's prediction as proposed in IRON-MAN. In our experiments, when the number of image frames in the frame window was selected to be more than 7, the prediction for the category became equal to zero for all the classes and hence, a frame window of more than 7 image frames could not be chosen for accurate predictions. Now, if we consider $t$ seconds as the time interval between two image frames in the video then because of Lemma 1 and Theorem 1 using IRON-MAN methodology we are only able to analyze a snippet of the video of $t \times 7$ seconds duration. If we consider the traffic categorization problem where each image frames were taken every 1.3 seconds (see Sec. \ref{dataset}) interval then using IRON-MAN we are able to analyze 9.1 (1.3 $\times$ 7) seconds of the video with high prediction accuracy.

In order to overcome the aforementioned limitations of IRON-MAN as mentioned earlier, we need to develop a more robust algorithm so that we are able to analyze larger snippets of videos (larger than 9.1 seconds for traffic categorization problem) without motion features more accurately.

	\section{Conclusion}\label{conclusion}
	In this paper, we propose IRON-MAN, which is capable of providing \textit{Temporal Motionless Analysis of Videos} ($TMAV$) i.e. analyzing videos without motion features and providing a holistic temporal analysis while utilizing predictions of the past image frames into consideration. Our approach is able to achieve 100\% prediction accuracy in analyzing video from the image frames for certain applications and it was also noticed that for our chosen applications VGG16 had better energy-efficiency on Exynos 5422 platform compared to ResNet50 using \textit{Energy Consumption per Training Image} ($ECTI$) metric comparison. Based on the results we have also shown that training CNN based approaches on MPSoCs could lead up to 4.8$\times$ (approx.) reduction in lifespan of the embedded device and therefore, it is recommended to perform the training off the embedded device for improved longevity. It is also shown that for traffic categorization application, our approach outperforms the state-of-the-art.
	
	{\small
		\bibliographystyle{IEEEtran}
		\bibliography{research_short}
	}
	
\end{document}